\documentclass[10pt,journal,compsoc]{IEEEtran}
\usepackage[switch]{lineno}

\ifCLASSOPTIONcompsoc
  \usepackage[nocompress]{cite}
\else
  \usepackage{cite}
\fi
\ifCLASSINFOpdf
\else
\fi
\usepackage{amsmath}
\usepackage{algorithmic}

\usepackage{array}

\usepackage{stfloats}
\usepackage{url}

\usepackage{verbatim}
\usepackage{booktabs} 
\usepackage{multirow}
\usepackage{amssymb}
\usepackage{hyperref}

\usepackage{graphicx}
 
\begin{document}
%
\title{Towards Medical Artificial General Intelligence via Knowledge-Enhanced Multimodal Pretraining}
%
%
%
%

\author{Bingqian~Lin$^{1}$\IEEEauthorrefmark{1}, Zicong Chen$^{1}$\IEEEauthorrefmark{1}, Mingjie Li$^{3}$\IEEEauthorrefmark{1}, Haokun Lin$^{1}$, Hang Xu$^{4}$, Yi Zhu$^{4}$, Jianzhuang Liu$^{4}$, \\Wenjia Cai$^{5}$, Lei Yang$^{6}$, Shen Zhao$^{1}$\IEEEauthorrefmark{2}, Chenfei Wu$^{7}$, Ling Chen$^{3}$, \\Xiaojun Chang$^{3}$, Yi Yang$^{8}$, Lei Xing$^{9}$, Xiaodan Liang$^{1,2}$\IEEEauthorrefmark{2}
	\IEEEcompsocitemizethanks{
	\IEEEcompsocthanksitem 
	\IEEEauthorrefmark{1}Equal Contribution\protect\\
	\IEEEcompsocthanksitem 
	\IEEEauthorrefmark{2}Corresponding Author\protect\\	\IEEEcompsocthanksitem
 $^1$Shenzhen Campus of Sun Yat-sen University. 
 $^2$Peng Cheng Laboratory.
 $^3$ReLER, AAII, University of Technology Sydney. $^4$Huawei Noah's Ark Lab. $^5$State Key Laboratory of Ophthalmology, Zhongshan Ophthalmic Center, Sun Yat-sen University, Guangdong Provincial Key Laboratory of Ophthalmology and Vision Science, Guangdong Provincial Clinical Research Center for Ocular Diseases, Guangzhou, China. $^6$The Department of Thoracic Surgery, The First Affiliated Hospital of Sun Yat-sen University. $^7$Sun Yat-sen University Cancer Center. $^8$Zhejiang University. $^9$Stanford University.
	}
		}

\IEEEtitleabstractindextext{%
\begin{abstract}
 Medical artificial general intelligence (MAGI) enables one foundation model to solve different medical tasks, which is very practical in the medical domain. It can significantly reduce the requirement of large amounts of task-specific data by sufficiently sharing medical knowledge among different tasks. However, due to the challenges of designing strongly generalizable models with limited and complex medical data, most existing approaches tend to develop task-specific models. To take a step towards MAGI, we propose a new paradigm called Medical-knOwledge-enhanced mulTimOdal pretRaining (MOTOR). In MOTOR, we combine two kinds of basic medical knowledge, i.e., general and specific knowledge, in a complementary manner to boost the general pretraining process. As a result, the foundation model with comprehensive basic knowledge can learn compact representations from pretraining radiographic data for better cross-modal alignment, i.e., matching the medical data of different modalities with the same semantic meanings. 
 MOTOR unifies the understanding and generation, which are two kinds of core intelligence of an AI system, into a single medical foundation model, to flexibly handle more diverse medical tasks.
 To enable a comprehensive evaluation and facilitate future research, we construct a medical multimodal benchmark including a wide range of downstream tasks, such as chest x-ray report generation and medical visual question answering. Extensive experiments on our benchmark show that MOTOR obtains promising results through simple task-oriented adaptation. The visualization shows that the injected  knowledge successfully highlights key information in the medical data, demonstrating the excellent interpretability of MOTOR. 
 Our MOTOR successfully mimics the human practice of fulfilling a ``medical student''  to accelerate the process of becoming a ``specialist''.
 We believe that our work makes a significant stride in realizing MAGI. 

\end{abstract}

}

\maketitle

\IEEEdisplaynontitleabstractindextext

%
\IEEEpeerreviewmaketitle

\IEEEraisesectionheading{\section*{Introduction}\label{sec:introduction}}
Developing medical artificial general intelligence (MAGI) is a practical and promising way to improve the efficiency for dealing with plenty of medical tasks. MAGI correlates medical data of different modalities (e.g., radiology images and reports) and deals with diverse medical tasks with one single foundation model. It facilitates medical knowledge sharing among different tasks and therefore largely reduces the requirement of collecting large-scale task-specific labelled data, which is usually difficult in the medical task.

Although MAGI is desired in the medical domain, building a strongly generalizable foundation model for realizing MAGI is quite challenging. Firstly, training the foundation model requires a vast amount of data, while accessing large-scale, diverse medical datasets is usually difficult. Secondly, the complexity of the medical data imposes great challenges on designing effective medical foundation models. Thirdly, high-performance computing devices are costly and not generally available for training a large foundation model.
Therefore, most previous approaches tend to design task-specific medical AI models~\cite{Chen2020GeneratingRR,liu2021exploring,Nguyen2019OvercomingDL,Zhan2020MedicalVQ}. 
For example, a chest X-ray interpretation model may be trained on a dataset in which every image has been explicitly labelled as positive or negative for pneumonia. As a result, the model would only detect pneumonia and would not be able to carry out other diagnostic exercises. 
Therefore, task-specific approaches may produce inflexible models. Moreover, the performance will be largely harmed when only a small amount of  task-specific data is available.

Recently, multimodal pretraining, especially the vision-language pretraining~\cite{tan2019lxmert,Zhang2021VinVLRV,li2020unicoder,li2020oscar,chen2020uniter,Li2021AlignBF,Li2022BLIPBL}, has disrupted task-specific paradigms and  shown great potential in realizing artificial general intelligence (AGI)~\cite{Fei2021TowardsAG}.
It greatly benefits from
the availability of large-scale image-text web data and the development of high-performance computing devices.
In vision-language pretraining, a foundation model is firstly pretrained on large-scale multimodal data to learn the general cross-modal alignment, i.e., matching the image-text pair with the same semantics. Then, it is generalized to various downstream tasks through task-oriented model architecture modification and finetuning. 
Inspired by the great progress of the multimodal pretraining made towards AGI, a natural idea is to transfer the multi-modal pretraining paradigm to realize MAGI. Nevertheless, different from natural images and captions, there exists severe redundancy in the medical multi-modal data. 
For example, descriptions of abnormality usually account for a small part of a radiology report. 
Therefore, direct transfer of the existing multi-modal pretraining paradigm developed in the natural image domain into the medical domain is not an effective way. 

\begin{figure*}[ht]
\begin{centering}
\includegraphics[width=0.98\linewidth]{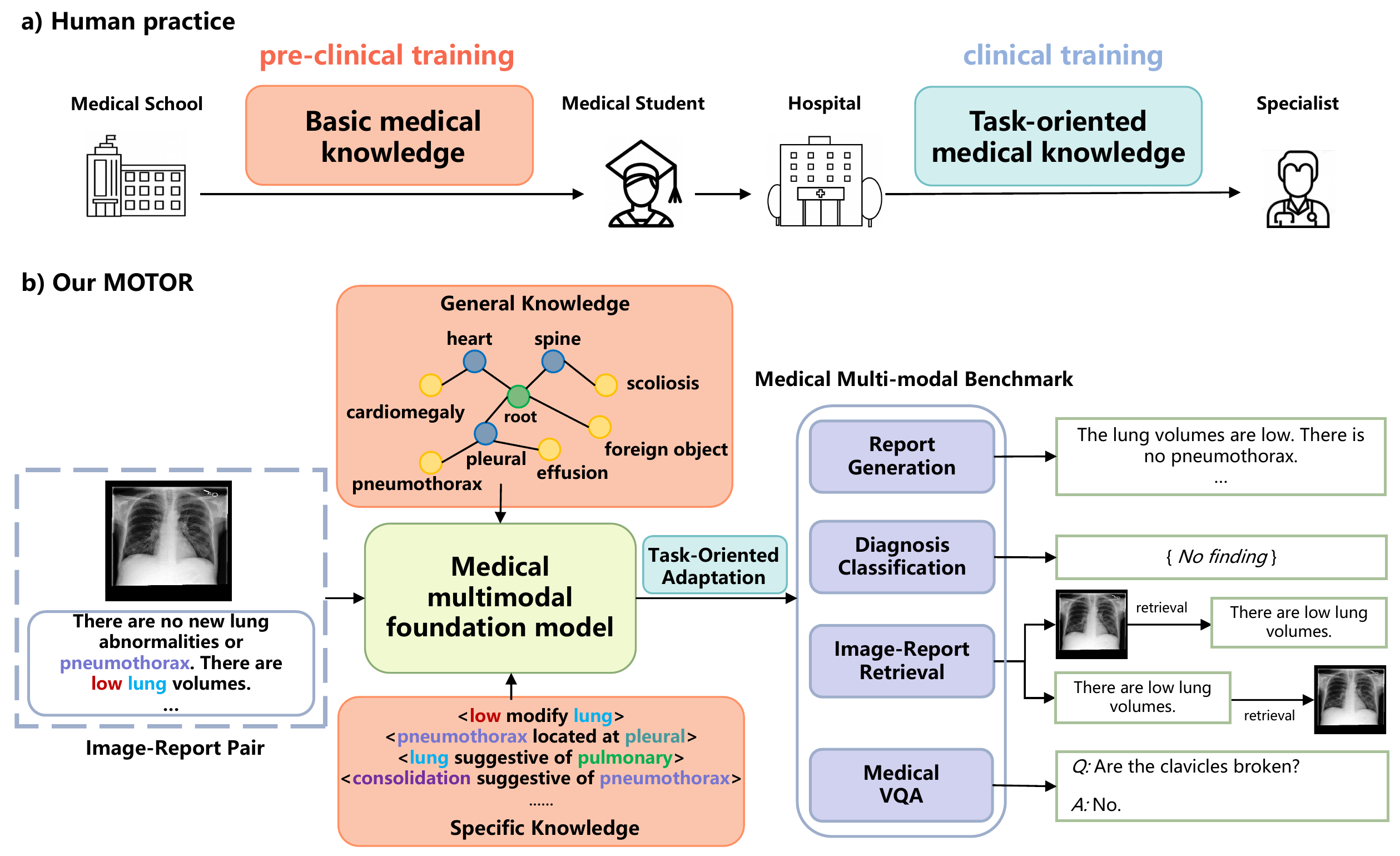}
\par\end{centering}
\caption{\label{fig:motivation}
Overview of MOTOR. In a) human practice, a specialist can be easily transferred from a medical student who owns basic medical knowledge through task-oriented adaptation. In b) our MOTOR, we follow this practice to build a foundation model to be a ``medical student'', by injecting both general knowledge and specific knowledge as the basic medical knowledge into the foundation model during pretraining. The general knowledge captures the hierarchy of common organs and diseases, and the specific knowledge contains semantic relations of them. Then through simple task-oriented adaptation, MOTOR shows strong generalization ability on a wide range of downstream medical tasks.
Due to the space limit, we only show part of the general knowledge in the figure. The complete architecture of the injected general knowledge is shown in Figure~\ref{fig:retrieval_gk}(a).
}
\vspace{-0.4cm}
\end{figure*}

Some recent works have attempted to introduce the multimodal pretraining into the medical domain~\cite{Chen2022MultimodalMA,Moon2021MultiModalUA,Chen2022AlignRA} by resorting to existing available large-scale medical multi-modal data. 
However, these methods captured a lot of wrong modality correspondence knowledge, since they directly learned modality alignment from redundant medical data.
Chen et al.~\cite{Chen2022MultimodalMA} proposed multi-modal masked autoencoders to perform medical vision-and-language pretraining through reconstructing missing pixels and tokens from randomly masked medical images and texts. 
Moon et al.~\cite{Moon2021MultiModalUA} adopted a BERT-based architecture  with a novel multi-modal attention masking scheme.
As a result, these methods still show limited generalization ability to downstream medical tasks.

In this paper,  to make a solid step towards MAGI, we propose a new paradigm, called Medical-knOwledge-enhanced multTimOdal pretRaining (MOTOR). 
MOTOR advances existing works by proposing a new knowledge-enhancing medical multi-modal pretraining paradigm for building a medical foundation model which is capable of both understanding and generation.  
In MOTOR, we introduce two kinds of basic medical knowledge, i.e., general and specific knowledge into pretraining in a complementary way, as shown in Figure~\ref{fig:motivation}b). Specifically, the general knowledge is instance-agnostic and helps the model learn hierarchical correlations between common organs and diseases.
Enhanced by the general knowledge, MOTOR is further injected with instance-related specific knowledge to capture semantic relations among organs and diseases. 
We further develop a text-based multi-label classification pretraining task in order to bridge the injection of the two kinds of knowledge and ensure the quality of the injected knowledge. 
As a result, the foundation model  owns sufficient basic medical knowledge and can learn correct modality correspondence from medical pretraining data which suffers from severe redundancy. 
Then, through simple  modification for the model architecture and finetuning with limited task data, MOTOR can be quickly adapted to a wide range of downstream medical tasks. 
Our MOTOR makes a successful attempt in realizing MAGI by imitating the real-world experience, that is, MOTOR  fulfills a  ``medical student'' who owns basic medical knowledge to understand general medical data, and therefore accelerates the process of becoming a ``specialist'' for solving new medical tasks (see Figure~\ref{fig:motivation}a)).

\begin{figure*}[t]
\begin{centering}
\includegraphics[width=0.98\linewidth]{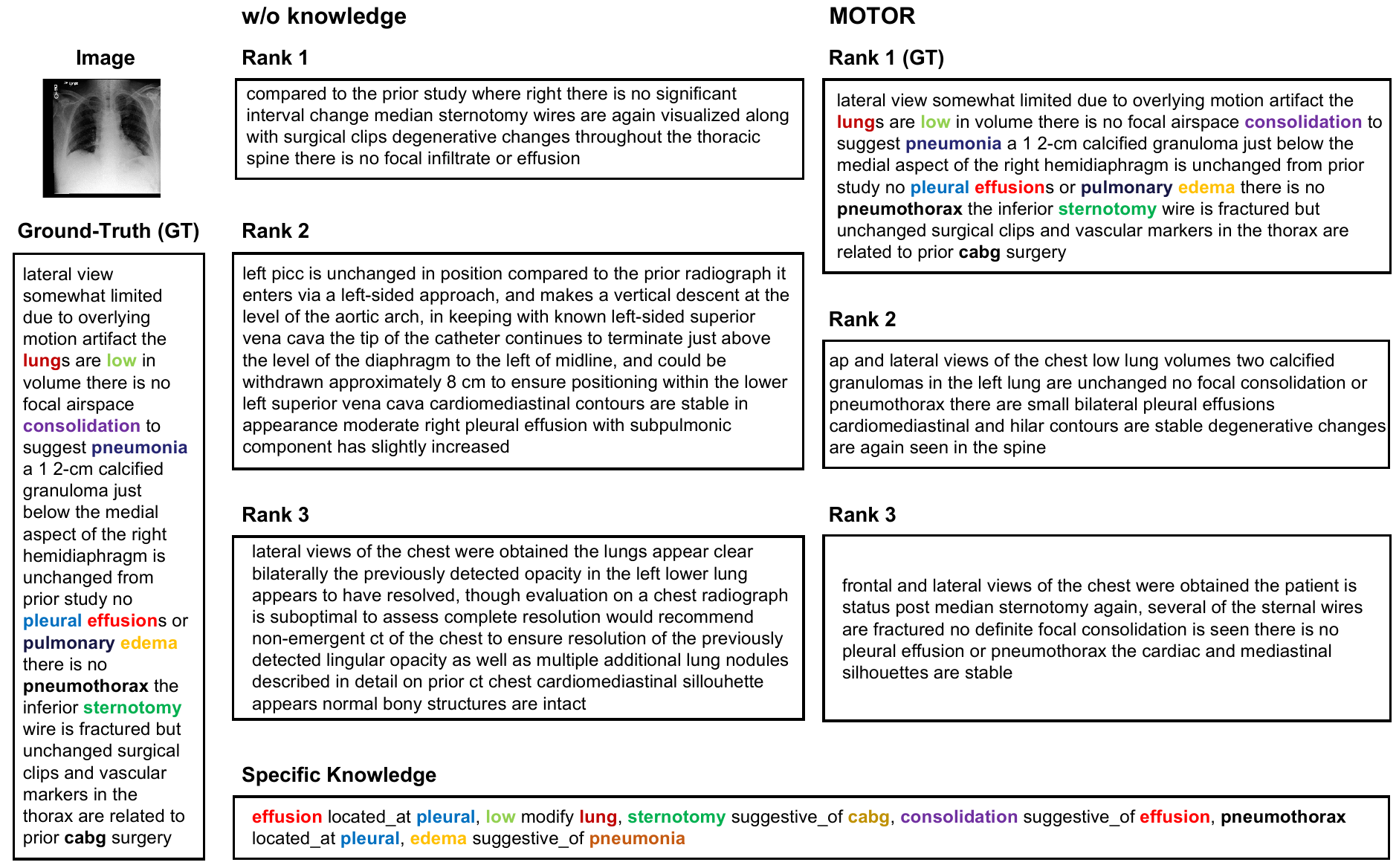}
\par\end{centering}
\caption{\label{fig:retrieval_sk} A comparison example between MOTOR and its non-knowledge variant (w/o knowledge) on image-report retrieval (IRR). The top 3 retrieved reports of both two methods using the given image are presented. The instance-related specific knowledge retrieved is given. The clinical entities recognized by Stanza~\cite{Qi2020StanzaAP} in the ground-truth (GT) report are manually highlighted with different color. 
}
\vspace{-0.2cm}
\end{figure*}

A recent work proposed by Chen et al.~\cite{Chen2022AlignRA} is closely related to our MOTOR. In~\cite{Chen2022AlignRA}, structured medical knowledge regarding to the entities in medical texts is also introduced during pretraining.
However, there exist two main differences between MOTOR and~\cite{Chen2022AlignRA}: 
1) Rather than adopting the pure knowledge graph in pretraining, MOTOR develops a novel knowledge-enhanced pretraining paradigm, where different kinds of knowledge are combined in a complementary manner. This significantly encourages the learning of correct modality correspondence from redundant medical multi-modal data. 
2) MOTOR enables a single medical foundation model to address both understanding-based and generation-based medical tasks, while~\cite{Chen2022AlignRA} leaves the generation tasks unresolved. 

To enable a comprehensive evaluation of the pretrained model and facilitate future research, we construct a medical multimodal benchmark including both understanding and generation tasks: Medical Report Generation, Image-Report Retrieval, Diagnosis Classification, and Medical Visual Question Answering. 
Experimental results  show that MOTOR outperforms (or is comparable to) previous state-of-the-art approaches on different tasks through simple task-oriented model modification and finetuning.
We compare MOTOR with its variant which pretrains without injecting the basic knowledge (``w/o knowledge'' for short) on multiple downstream tasks. 
The significant performance improvement of MOTOR over its ``w/o knowledge'' variant demonstrates the effectiveness of the proposed knowledge-enhancing pretraining paradigm.
We conduct extensive visualization with human evaluation by our expert radiologists.
The visualization results show that the injected general and specific knowledge effectively highlights key information in the redundant medical multi-modal data to enhance the modality alignment, demonstrating the excellent interpretability of MOTOR.
We believe our work will be a meaningful reference to encourage  future research in realizing MAGI.

\begin{table*}
\vspace{-0.2cm}
    \caption{Image-report retrieval (IRR) performance comparison under zero-shot and finetuning settings on MIMIC-CXR~\cite{Johnson2019MIMICCXRAL}. GK and SK mean general knowledge and specific knowledge, respectively.}

    \small
    \centering
    \resizebox{\linewidth}{!}{
    
    \begin{tabular}{l|cccccc|cccccc}
    \specialrule{.1em}{.05em}{.05em}
            
			\multirow{2}{*}{Method}&\multicolumn{6}{c|}{Zero-shot}&\multicolumn{6}{c}{Finetuning}\cr\cline{2-13}
            &\multicolumn{3}{c}{RR}&\multicolumn{3}{c|}{IR}&\multicolumn{3}{c}{RR}&\multicolumn{3}{c}{IR}\cr\cline{2-13}
            &R@1&R@5&R@10&R@1&R@5&R@10&R@1&R@5&R@10&R@1&R@5&R@10\\
            
            \hline
            MOTOR w/o knowledge&9.05&26.41&36.78&9.75&28.54&38.49&10.58&31.60&42.61&\textbf{12.05}&32.87&42.61\\
            MOTOR w GK&10.26&30.61&41.73&10.65&30.04&42.72&10.01&31.34&42.38&10.55&31.26&43.47\\
            MOTOR w SK&9.36&29.70&40.49&10.94&29.86&41.45&10.42&31.73&42.77&10.89&30.48&42.43\\
             MOTOR w GK+SK&10.81&\textbf{31.44}&42.76&11.09&31.89&42.90&9.67&30.12&42.61&11.04&30.84&41.91\\
             MOTOR (ours)&\textbf{10.83}&31.00&\textbf{42.87}&\textbf{11.35}&\textbf{31.99}&\textbf{43.26}&\textbf{10.96}&\textbf{31.93}&\textbf{42.90}&12.00&\textbf{33.10}&\textbf{44.32}\\
        
 \specialrule{.1em}{.05em}{.05em}
    \end{tabular}
}
\label{tab:retrieval}
\end{table*}

\section*{Results}
Following~\cite{Moon2021MultiModalUA}, we also use the public large-scale medical multi-modal dataset MIMIC-CXR~\cite{Johnson2019MIMICCXRAL}  for pretraining, which contains plenty of radiology Chest X-ray images and corresponding free-text reports.
We include multiple popular radiology-based datasets, i.e., MIMIC-CXR~\cite{Johnson2019MIMICCXRAL}, IU-Xray~\cite{DemnerFushman2016PreparingAC}, ChestX-ray 14~\cite{Wang2017ChestXRay8HC}, VQA-RAD~\cite{Lau2018ADO}, and SLAKE~\cite{Liu2021SlakeAS}, into our constructed multi-modal benchmark. Among them, VQA-RAD and SLAKE contain the radiology images regarding the organs of the whole body, which can effectively verify the generalization ability of the model to different organs. 
In the following, we present the experimental results with detailed analyses on the downstream medical tasks in our benchmark, including image-report retrieval, medical report generation, diagnosis classification, and medical visual question answering.

\begin{figure*}[t]
\begin{centering}
\includegraphics[width=0.98\linewidth]{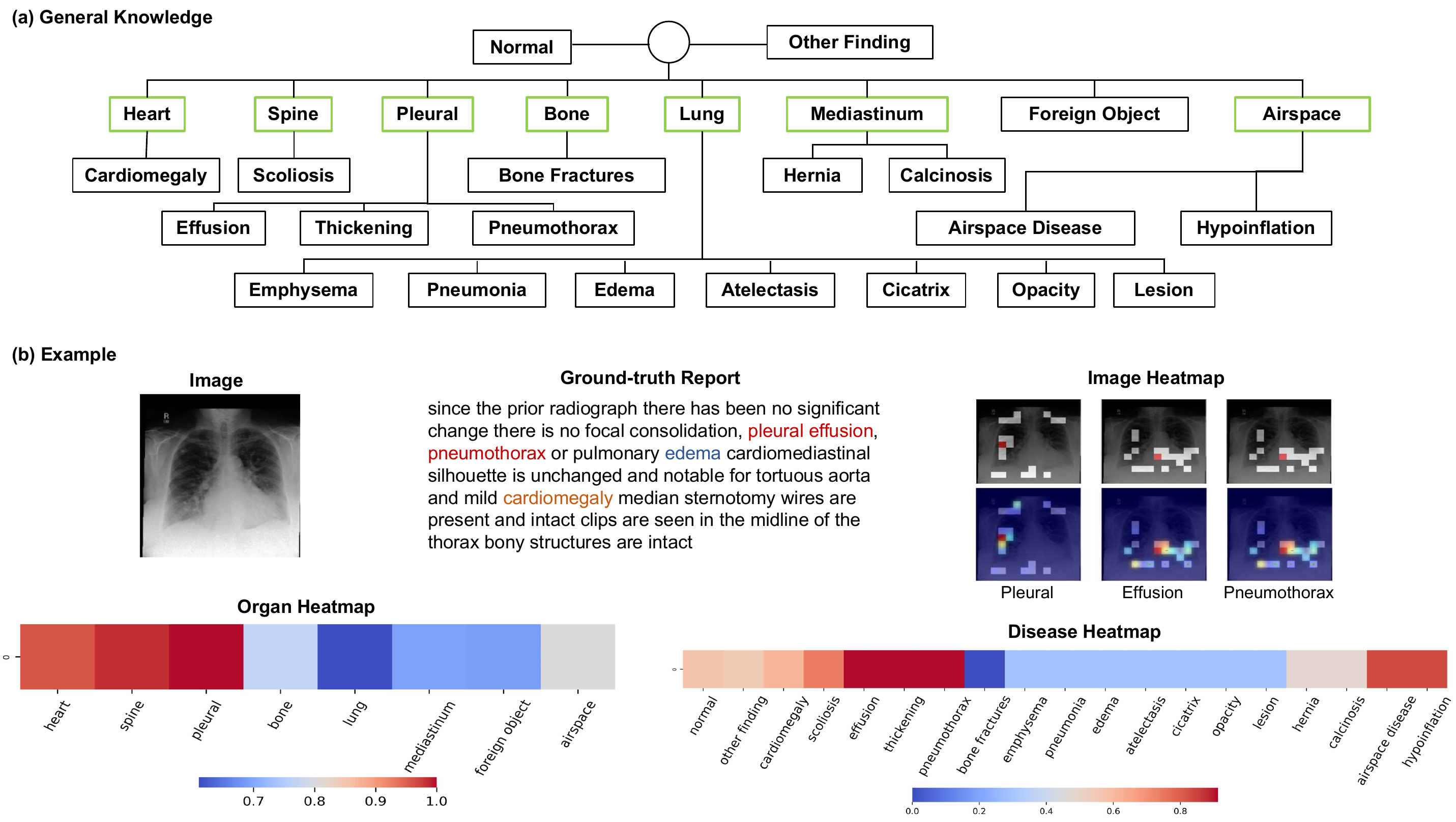}
\par\end{centering}
\caption{\label{fig:retrieval_gk} (a) Visualization of the injected general knowledge. The organ nodes are annotated by green boxes. (b)Visualization of image-to-node attention heatmaps (organ heatmap \& disease heatmap) and node-to-image attention heatmaps (image heatmap). The visualized nodes are extracted from the general knowledge.
}
\vspace{-0.2cm}
\end{figure*}

\begin{figure*}[t]
\begin{centering}
\includegraphics[width=0.98\linewidth]{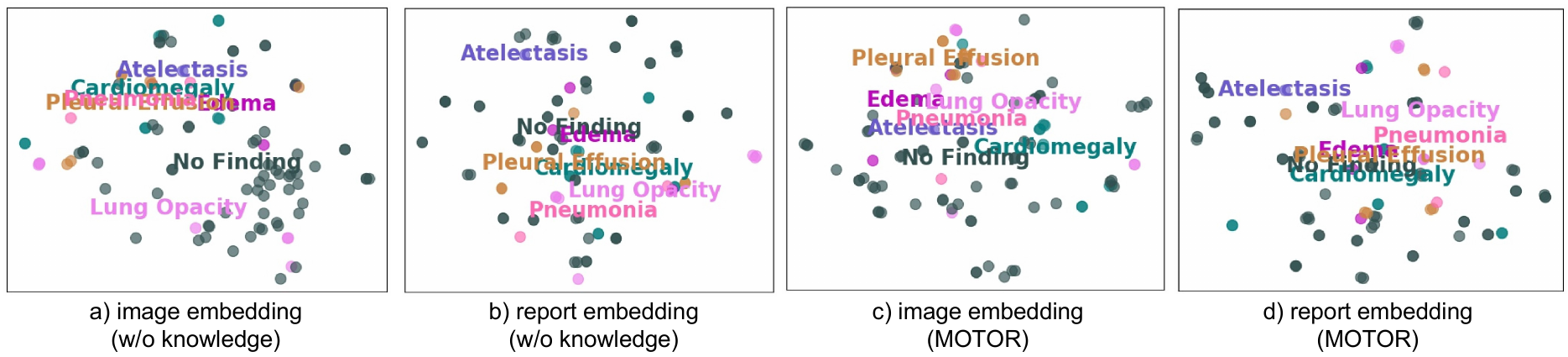}
\par\end{centering}
\caption{\label{fig:vis_emb} Visualization of multi-modal embedding distributions of MOTOR and its non-knowledge variant (w/o knowledge) on image-report retrieval (IRR) task. In our injected general knowledge, ``edema'', ``lung opacity'', ``pneumonia'', and ``atelectasis'' belong to the organ ``lung''. ``Cardiomegaly'' belongs to the organ ``heart''. ``Pleural effusion'' belongs to the organ ``pleural''.
}
\vspace{-0.2cm}
\end{figure*}

\begin{figure*}[t]
\begin{centering}
\includegraphics[width=0.98\linewidth]{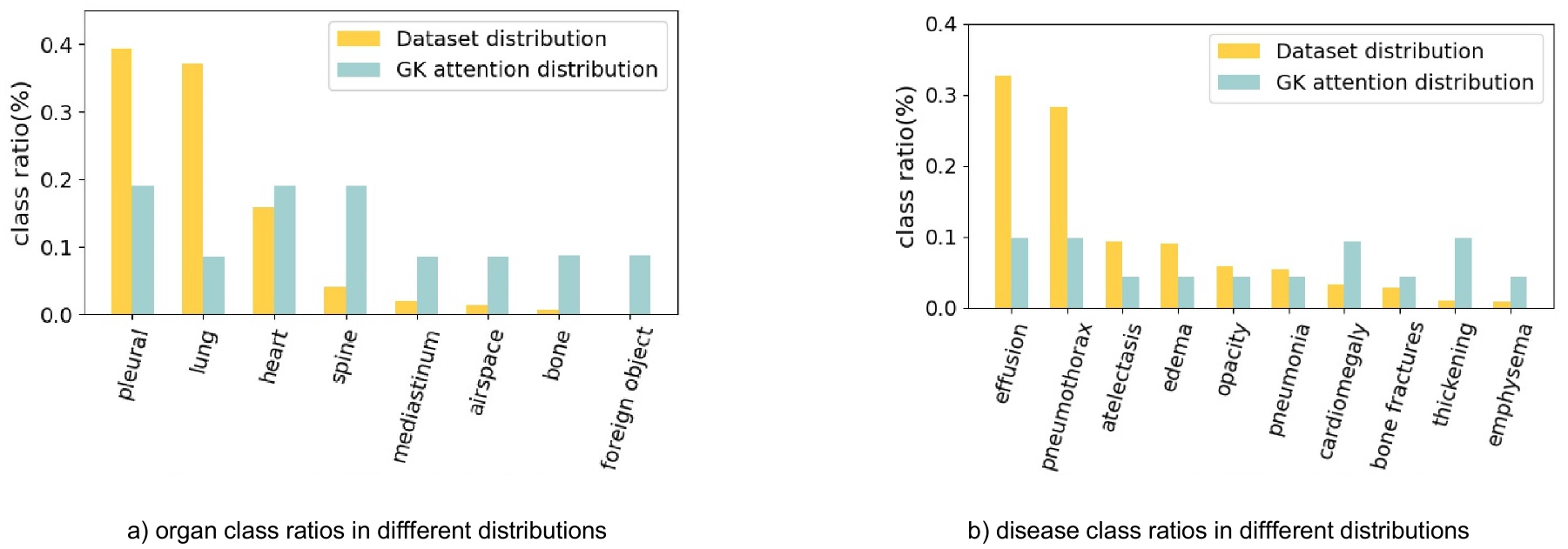}
\par\end{centering}
\caption{\label{fig:vis_dis} Visualization of class ratios in distributions of the pretraining dataset and the general knowledge (GK) attention. For obtaining the GK attention, we calculate the average value of the image-to-GK attention for each image in the pretraining dataset.  
}
\vspace{-0.4cm}
\end{figure*}

\noindent\textbf{Image-report retrieval.}
For
multi-modal pretraining in the natural image domain~\cite{Li2022BLIPBL,Li2021AlignBF,Zhang2021VinVLRV}, the image-text retrieval task is a simple and effective way to evaluate the cross-modal understanding ability of pretrained models. Therefore, we develop an  image-report retrieval (IRR) task in our benchmark.
Two subtasks are included in IRR: report retrieval (RR), where images and reports are queries and targets, respectively; and image retrieval (IR), where reports are queries and images are targets. The R@K (recall with top $k$ predictions) metric is used for the performance evaluation for both subtasks. We directly use the pretraining dataset MIMIC-CXR~\cite{Johnson2019MIMICCXRAL} for evaluating the IRR performance under both zero-shot and finetuning settings.

Since there are no available reported results of existing methods on the IRR task on MIMIC-CXR~\cite{Johnson2019MIMICCXRAL}, we compare MOTOR and its variants for analyzing the impact of different components in MOTOR deeply. We present the results in Table~\ref{tab:retrieval}.
Table~\ref{tab:retrieval} comprehensively show the effectiveness of each method component of MOTOR. Moreover, the significant performance superiority of MOTOR over other variants under the zero-shot setting sufficiently demonstrates the excellent generalization ability of MOTOR. Concretely, under the zero-shot setting, ``MOTOR w GK/SK'' outperforms ``MOTOR w/o knowledge'' by a large margin.  
This demonstrates that both the general and specific knowledge can improve the medical multi-modal pretraining. Moreover, the performance improvement of  ``MOTOR w GK/SK'' over  ``MOTOR w GK+SK''  reveals that the general and specific knowledge can complement each other well. From both zero-shot and finetuning settings, we can also observe that our full model (MOTOR (ours)), which introduces the two kinds of knowledge with the text-based multi-label classification pretraining objective, achieves the best performance among all variants.

Figure~\ref{fig:retrieval_sk} gives a comparison example between MOTOR and its non-knowledge variant (w/o knowledge) on IRR. The injected instance-related specific knowledge is also given in Figure~\ref{fig:retrieval_sk}. From Figure~\ref{fig:retrieval_sk}, we can observe that with the help of knowledge, MOTOR owns better image-report retrieval capability than the ``w/o knowledge'' variant. Moreover, according to the evaluation of expert radiologists, our injected specific knowledge can enable the model to learn the complicated correspondence between paired image-report data, and therefore leads to successful retrieval. Specifically, MOTOR successfully retrieves the GT report at rank 1 while “w/o knowledge” fails to retrieve it in its top 3 predictions. We can find from Figure~\ref{fig:retrieval_sk} that the specific knowledge indicates many crucial clinical entities matching the ground-truth report, such as “consolidation” and “pneumonia”. Beyond the entities, the specific knowledge also contains rich semantic relations among these entities, such as “sternotomy suggestive of cabg” and “consolidation suggestive of effusion”, which effectively facilitates the deep alignment between medical image-report data.

Figure~\ref{fig:retrieval_gk} gives a visualization example of general knowledge attention. Specifically, we visualize image-to-node attention heatmaps (organ heatmap \& disease heatmap) where nodes are extracted from the general knowledge graph, which is shown in Figure~\ref{fig:retrieval_gk}(a). Moreover, we show the node-to-image attention heatmaps (image heatmap) for the nodes ``pleural'', ``effusion'', and ``pneumothorax'', which are mentioned in the ground-truth (GT) report and have the highest attention value in the image-to-node attention heatmaps. From organ/disease heatmaps in Figure~\ref{fig:retrieval_gk}, we can find that with the help of general knowledge, MOTOR can capture the hierarchy of organs and diseases. Concretely, the entities in the same branch in the general knowledge graph show consistent attention distributions. For example, both ``pleural'' and ``effusion'' show consistently high attention values. Moreover, we can also observe that MOTOR performs well in alignment between the medical image-text data. Firstly, through organ heatmaps, disease heatmaps and the GT report, we can find that MOTOR  successfully highlights crucial entities in the GT report, such as ``cardiomegaly'' and ``effusion'' through the image. 
Moreover, evaluated by expert radiologists for the image heatmaps, MOTOR can perform well in alignment between the organ/disease entities and the related image regions.

Figure~\ref{fig:vis_emb} presents the visualization of multi-modal embedding distributions of MOTOR and its non-knowledge variant (``w/o knowledge'') on image-report retrieval (IRR) task.  The distribution comparison in Figure~\ref{fig:vis_emb} shows that MOTOR can perform significantly better cross-modal alignment than the ``w/o knowledge'' variant. 
The visualization also reveals that the general knowledge can enable MOTOR to learn the relations between organs and diseases. Concretely, there exists a large difference between the distributions of image embeddings and the report embeddings of the ``w/o knowledge'' model. In contrast, such difference is much smaller in MOTOR as we can see in Figure~\ref{fig:vis_emb}c)$\sim$d). Moreover, we can also find that MOTOR better captures the relations among disease categories than the “w/o knowledge” variant.  For example, in our injected general knowledge, “edema”, “pneumonia”, “lung opacity”, and “atelectasis” belong to the organ “lung”. By comparing the embedding distributions of MOTOR and the “w/o knowledge”  variant, we can find that the embeddings of these disease categories are much closer in MOTOR than in the “w/o knowledge” variant. 

Figure~\ref{fig:vis_dis} gives the visualization of category  distributions of the pretraining dataset and the general knowledge attention. 
From Figure~\ref{fig:vis_dis}, we can find that the  
class distributions of both organ and disease show good consistency in pretraining dataset and the general knowledge attention. This shows that with the help of the general knowledge, MOTOR can capture the basic class distribution of different organs and diseases from the general medical dataset. For example, in Figure~\ref{fig:vis_dis}a), ``pleural'' and ``heart'' have relatively large class ratios while ``airspace'' and ``bone'' have relative small class ratios in both dataset distribution and GK attention distribution. In Figure~\ref{fig:vis_dis}b), ``effusion'' and ``pneumothorax'' have relative large class ratios while ``opacity'' and ``pneumonia'' have relative small class ratios in both dataset distribution and GK attention distribution. By sufficiently understanding general medical data, MOTOR can be able to generalize better to downstream medical tasks.

\begin{table*}
\centering

\caption{Medical report generation performance of different methods on IU-Xray~\cite{DemnerFushman2016PreparingAC}. * represents the main metric. GK and SK represent general knowledge and specific knowledge, respectively.}
\begin{tabular}{l|cccc} 
\toprule
Methods    & CIDEr* & ROUGE\_L & METEOR & BLEU\_4  \\ 
\hline
CoAtt~\cite{Jing2018OnTA}        & 0.277 & 0.369    & -         & 0.154    \\
R2Gen~\cite{Chen2020GeneratingRR}        & 0.398 & 0.322    & 0.165     & 0.124    \\
Con-Trans~\cite{alfarghaly2021automated}  & 0.257 & 0.289    & 0.164    & 0.111    \\
SentSAT+KG~\cite{Zhang2020WhenRR} & 0.304 & 0.367    & -         & 0.147    \\
PPKED~\cite{liu2021exploring}       & 0.351 & \textbf{0.376}    & 0.190     & \textbf{0.168}    \\ 
\hline
MOTOR w/o knowledge &0.386&0.278&0.164&0.113\\
MOTOR w GK &    0.597&0.311&0.185&0.142    \\
MOTOR w SK &0.583&0.308&0.189&0.146\\
MOTOR w GK+SK &0.578&0.306&\textbf{0.194}&0.142\\
MOTOR (ours) &\textbf{0.699}&0.314&0.193&0.156\\
\bottomrule
\end{tabular}
\label{tab:generation}
\end{table*}

\begin{figure*}[t]
\begin{centering}
\includegraphics[width=0.98\linewidth]{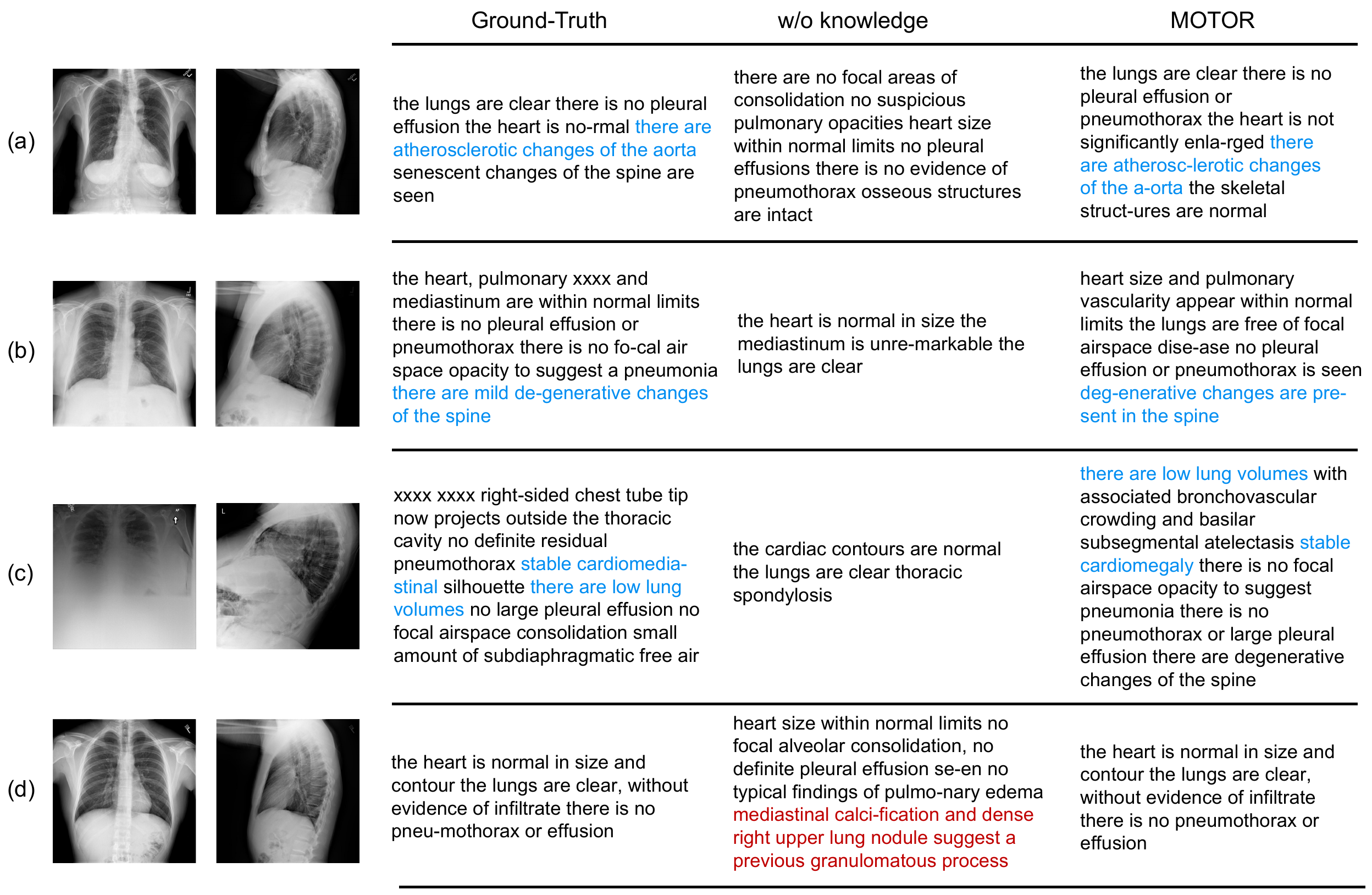}
\par\end{centering}
\caption{\label{fig:generated reports} Visualization examples of generated reports of MOTOR and its non-knowledge variant (w/o knowledge). Correct details matching the ground-truth reports are highlighted in blue. Wrong ones are highlighted in red.
}
\vspace{-0.4cm}
\end{figure*}

\noindent\textbf{Medical report generation.}
Medical report generation (MRG) is a medical cross-modal generation task, which requires the system to automatically generate a free-text report given a medical image. 
Our benchmark contains the chest x-ray report generation task, which is one of the most popular MRG tasks.
Following existing MRG works~\cite{Jing2018OnTA,Li2018HybridRR,Chen2020GeneratingRR,Zhang2020WhenRR,liu2021exploring}, we use the IU-Xray~\cite{DemnerFushman2016PreparingAC} dataset to verify the report generation performance of our model. Note that since MIMIC-CXR~\cite{Johnson2019MIMICCXRAL} is used for pretraining using the same objective with the MRG task, we do not compare MOTOR with existing methods on MIMIC-CXR for fairness. The widely-used BLEU~\cite{Papineni2002BleuAM}, METEOR~\cite{Banerjee2005METEORAA}, ROUGE-L~\cite{Lin2004ROUGEAP}, and CIDEr~\cite{Vedantam2015CIDErCI} for MRG are adopted as the evaluation metrics.

The performance comparison results of different methods are presented in Table~\ref{tab:generation}. From  Table~\ref{tab:generation}, we can find that MOTOR significantly outperforms state-of-the-art task-specific approaches in CIDEr~\cite{Vedantam2015CIDErCI}, which is an important metric in evaluating text generation systems. The excellent performance of CIDEr reveals that MOTOR tends to generate accurate topics rather than repeating the sentence in the training set simply. For other metrics, MOTOR also achieves comparable performance with existing task-specific methods. Moreover, the superiority of MOTOR over its variants shows the effectiveness of the proposed knowledge-enhanced pretraining paradigm. Specifically, we can find that consistent with  Table~\ref{tab:retrieval}, ``MOTOR w GK/SK'' also outperforms ``MOTOR w/o knowledge'' by a large margin. Note that MOTOR and its knowledge-based variants, such as ``MOTOR w GK'' and ``MOTOR w SK'', significantly improve CIDEr~\cite{Vedantam2015CIDErCI}. These results show the effectiveness of two kinds of injected knowledge in highlighting key information in the multi-modal medical data. Moreover, our full model is superior over ``MOTOR w GK/SK'' and ``MOTOR w GK+SK''. This demonstrates that general and specific knowledge effectively complement each other with the help of the text-based multi-label classification pretraining objective. 

We give some generated reports of MOTOR and the ``w/o knowledge'' variant in Figure~\ref{fig:generated reports} for analyzing the impact of knowledge deeply.  Figure~\ref{fig:generated reports}(a)-(c) are disease cases, and Figure~\ref{fig:generated reports}(d) is a normal case. 
According to the human evaluation by expert radiologists, the generated reports by MOTOR contain more accurate details about diseases than the ``w/o knowledge'' variant.
For example, in Figure~\ref{fig:generated reports}(a) and (b), the generated reports of MOTOR accurately indicate the ``atherosclerotic changes of the aorta'' and ``degenerative changes of the spine'', respectively. The ``w/o knowledge'' variant, however, misses these important disease details. Moreover, from the normal case shown in Figure~\ref{fig:generated reports}(d), we can observe that the generated reports of ``w/o knowledge'' variant contains many incorrect disease details which are unexpected. This shows that without the help of knowledge, the model may learn wrong modality correspondence from the complex and redundant image-report data.       


\begin{table*}
\centering
\caption{Diagnosis classification performance (AUROC scores) of different methods on ChestX-ray14 dataset. $^{\diamondsuit}$ and $^{\heartsuit}$ represents using ResNet50~\cite{He2016DeepRL} and DenseNet-121~\cite{Huang2017DenselyCC} as the backbone, respectively. Atel, Card, Effu, Infi, Mass, Nodu, Pne1, Pne2, Cons, Edem, Emph, Fibr, P.T., and Hern represent
Atelectasis, Cardiomegaly, Effusion,
Infiltration, Mass, Nodule, Pneumonia, Pneumothorax, Consolidation, Edema, Emphysema, Fibrosis, Pleural Thickening, and Hernia, respectively.
}
\resizebox{\linewidth}{!}{
\begin{tabular}{l|llllllllllllll|l} 
\toprule
Methods     & Atel  & Card  & Effu  & Infi  & Mass  & Nodu  & Pne1  & Pne2  & Cons  & Edem  & Emph  & Fibr  & P.T.  & Hern  & Mean            \\ 
\hline
\hline
Wang et al.$^{\diamondsuit}$~\cite{Wang2017ChestXRay8HC} & 0.700 & 0.810 & 0.759 & 0.661 & 0.693 & 0.669 & 0.658 & 0.799 & 0.703 & 0.805 & 0.833 & 0.786 & 0.684 & 0.872 & 0.745           \\
Guan et al.$^{\diamondsuit}$~\cite{guan2020multi} & 0.779 & 0.879 & 0.824 & 0.694 & 0.831 & 0.766 & 0.726 & 0.858 & \textbf{0.758} & 0.850 & 0.909 & \textbf{0.832} & 0.778 & 0.906 & 0.814           \\ 
\hline
\hline
Guan et al.$^{\heartsuit}$~\cite{guan2020multi} & 0.781 & 0.883 & 0.831 & 0.697 & 0.830 & 0.764 & 0.725 & 0.866 & 0.758 & 0.853 & 0.911 & 0.826 & 0.780 & \textbf{0.918} & 0.816           \\
Kim et al.$^{\heartsuit}$~\cite{kim2021xprotonet}  & 0.780 & 0.887 & 0.835 & 0.710 & 0.831 & \textbf{0.804} & \textbf{0.734} & \textbf{0.871} & 0.747 & 0.840 & \textbf{0.941} & 0.815 & \textbf{0.799} & 0.909 & \textbf{0.822}  \\ 
\hline
\hline
MOTOR w/o knowledge        &    0.722   &    0.859   &  0.761     &    0.659  & 0.749&   0.681    &  0.631     &   0.821    &    0.660   &    0.775   &   0.862    &  0.782     &      0.741   &   0.911    &    0.756
\\
MOTOR (ours)&\textbf{0.790}&\textbf{0.944}&\textbf{0.882}&\textbf{0.715}&\textbf{0.846}&0.768&0.714&0.850&0.728&\textbf{0.893}&0.883&0.798&0.762&0.908&0.820\\

\bottomrule
\end{tabular}}
\label{tab:chest 14}
\end{table*}

\begin{table*}
\centering
\caption{Diagnosis classification performance of different methods on MIMIC-CXR dataset. * represents the re-implementation.}
\begin{tabular}{l|c|c|c|c} 
\toprule
Methods       & Macro-F1 (\%)      & Micro-F1 (\%)&F1 (\%)&AUROC (\%)       \\ 
\hline
ResNet101~\cite{He2016DeepRL}*     & 23.1          & 40.8 &37.1&75.2\\
DenseNet121~\cite{Huang2017DenselyCC}*   & 25.0          & 40.5&36.8&76.4           \\
\hline
MOTOR w/o knowledge  & 25.6& 42.5&39.4&76.6 \\    MOTOR (ours)&\textbf{28.78}&\textbf{44.01}&\textbf{41.49}&\textbf{77.19}      \\

\bottomrule
\end{tabular}
\label{tab:mimic classification}
\vspace{-0.4cm}
\end{table*}

\noindent\textbf{Diagnosis classification.}
Our benchmark contains the diagnosis classification task,
which is formulated as a multi-label image classification task including 14 common radiographic observations.
We adopt MIMIC-CXR~\cite{Johnson2019MIMICCXRAL} and ChestX-ray14~\cite{Wang2017ChestXRay8HC} for diagnosis classification. For MIMIC-CXR, the 14 categories are consistent with those in~\cite{Irvin2019CheXpertAL}. The evaluation metrics are F1, macro-F1, micro-F1, and AUROC.

The results on ChestX-ray14~\cite{Wang2017ChestXRay8HC} and MIMIC-CXR~\cite{Johnson2019MIMICCXRAL} are given in Table~\ref{tab:chest 14} and Table~\ref{tab:mimic classification}, respectively. From Table~\ref{tab:chest 14}, we can find that MOTOR outperforms ``MOTOR w/o knowledge'' by a significant margin on the average classification performance and the classification performance regarding different disease categories, showing the effectiveness of the proposed knowledge-enhancing strategy. 
Moreover, MOTOR is comparable to the state-of-the-art method~\cite{kim2021xprotonet} with a simple ViT architecture followed by a multi-label classification head.
Since there exists a gap between the task goals of pretraining and  diagnosis classification, where the former aims at image-text alignment and the latter aims at multi-label classification, the cross-modal alignment ability of the pretrained model cannot be sufficiently utilized in this downstream task.
In future work, it is promising to modify the training goal of the diagnosis classification task to image-text alignment like~\cite{Radford2021LearningTV,Zhou2021LearningTP,Zhou2022ConditionalPL} developed in the natural image domain, to better adapt the pretrained medical multi-modal model to the diagnosis classification task. 

Table~\ref{tab:mimic classification} shows consistent results with Table~\ref{tab:chest 14}, that is, MOTOR is superior over ``MOTOR w/o knowledge'' in all used metrics. Moreover, we can find that MOTOR outperforms ResNet101~\cite{He2016DeepRL} and DenseNet121~\cite{huang2017densely} at a large margin in most metrics, such as Macro-F1 and Micro-F1, showing that with the help of the proposed knowledge-enhanced pretraining strategy, the pure ViT backbone can generate better encodings of the medical images.  


\begin{table*}
\vspace{-0.2cm}
    \caption{Medical visual question answering accuracy (\%) of different methods on VQA-RAD and SLAKE datasets.}

    \small
    \centering
    \resizebox{\linewidth}{!}{
    
    \begin{tabular}{l|cccccc}
    \specialrule{.1em}{.05em}{.05em}
			\multirow{2}{*}{Method}&\multicolumn{3}{c}{VQA-RAD}&\multicolumn{3}{c}{SLAKE}\cr\cline{2-7}
            &Open-ended&Closed-ended&Overall&Open-ended&Closed-ended&Overall\\
            
            \hline
            SAN~\cite{Yang2016StackedAN}&31.30&69.50&54.30&74.00&79.10&76.00\\
            BAN~\cite{Kim2018BilinearAN}&37.40&72.10&58.30&74.60&79.10&76.30\\
            MEVF-SAN~\cite{Nguyen2019OvercomingDL}&49.20&73.90&64.10&75.30&78.40&76.50\\
            MEVF-BAN~\cite{Nguyen2019OvercomingDL}&49.20&77.20&66.10&77.80&79.80&78.60\\
            QCR+TCR~\cite{Zhan2020MedicalVQ}&60.00&\textbf{79.30}&\textbf{71.60}&-&-&-\\
            \hline
            MOTOR w/o knowledge&62.57&73.90&69.41&78.14&83.65&80.30\\
            MOTOR (ours)&\textbf{64.80}&74.63&70.73&\textbf{81.09}&\textbf{84.13}&\textbf{82.28}\\
        
 \specialrule{.1em}{.05em}{.05em}
    \end{tabular}
}
\label{tab:vqa}
\end{table*}

\begin{figure*}[t]
\begin{centering}
\includegraphics[width=0.98\linewidth]{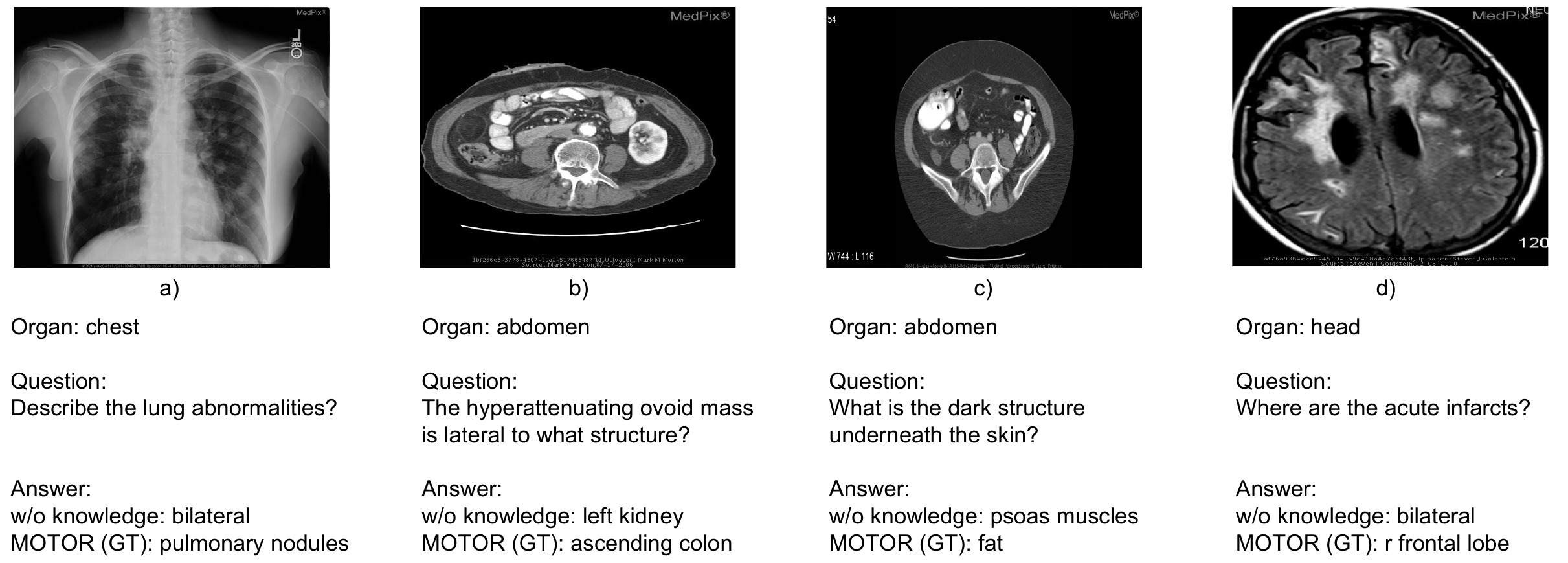}
\par\end{centering}
\caption{\label{fig:vqa} Visualization examples of MOTOR and its non-knowledge variant (w/o knowledge) on VQA-RAD~\cite{Lau2018ADO} dataset. The examples include different body parts with open-ended questions. For each example, MOTOR's answer is the same as the ground-truth (GT) answer. 
}
\vspace{-0.4cm}
\end{figure*}

\noindent\textbf{Medical visual question answering.} Medical visual question answering (Med-VQA) task takes a medical image and a clinical question about the image as input and asks the model to output an answer in natural language. We include two publicly available Med-VQA datasets, i.e., VQA-RAD~\cite{Lau2018ADO} and SLAKE~\cite{Liu2021SlakeAS}, in our benchmark. 
Both VQA-RAD and SLAKE contain images of different organs. According to the answer form, the questions in both VQA-RAD and SLAKE can be categorized into two types. The answers to closed-ended questions are ``yes/no'' or other limited choices and the answers to open-ended questions are free-form texts. Therefore, answering open-ended questions is much harder than answering closed-ended ones since it requires more challenging cross-modal alignment. Following existing Med-VQA works~\cite{Zhan2020MedicalVQ,Liu2021SlakeAS}, we use Accuracy as the evaluation metric.

The results on VQA-RAD and SLAKE are presented in Table~\ref{tab:vqa}. From Table~\ref{tab:vqa}, we can find that MOTOR outperforms ``MOTOR w/o knowledge'' in both VQA-RAD and SLAKE, demonstrating the effectiveness of the proposed knowledge-enhanced pretraining paradigm. Moreover, the performance gap between ``MOTOR w/o knowledge'' and MOTOR on open-ended questions is obviously larger than that on closed-ended questions, showing that MOTOR can capture more complicated correspondence between medical multi-modal image-text data with the help of the knowledge. Although there exists a big domain gap between the pretraining dataset (i.e., MIMIC-CXR) and the VQA dataset, we can find that MOTOR is still superior over (or comparable to) the state-of-the-art Med-VQA models. This shows the strong generalization ability of MOTOR.

Figure~\ref{fig:vqa} presents some visualization examples of MOTOR and its non-knowledge variant (w/o knowledge) on VQA-RAD~\cite{Lau2018ADO} dataset. From Figure~\ref{fig:vqa}, we can find that MOTOR has better multi-modal understanding and cross-modal alignment ability than the ``w/o knowledge'' variant, and shows superior performance on challenging open-ended questions. For example, in Figure~\ref{fig:vqa}a), for the question ``describe the lung abnormalities'', MOTOR successfully chooses the correct answer ``pulmonary nodules''. The ``w/o knowledge'' variant, however, cannot well understand the meaning of the question based on the given image. Moreover, we can also observe from the visualization results that MOTOR shows strong generalization abilities to different body parts, demonstrating that it is promising to transfer MOTOR to deal with diverse downstream medical tasks. 


\begin{figure*}[t]
\begin{centering}
\includegraphics[width=0.98\linewidth]{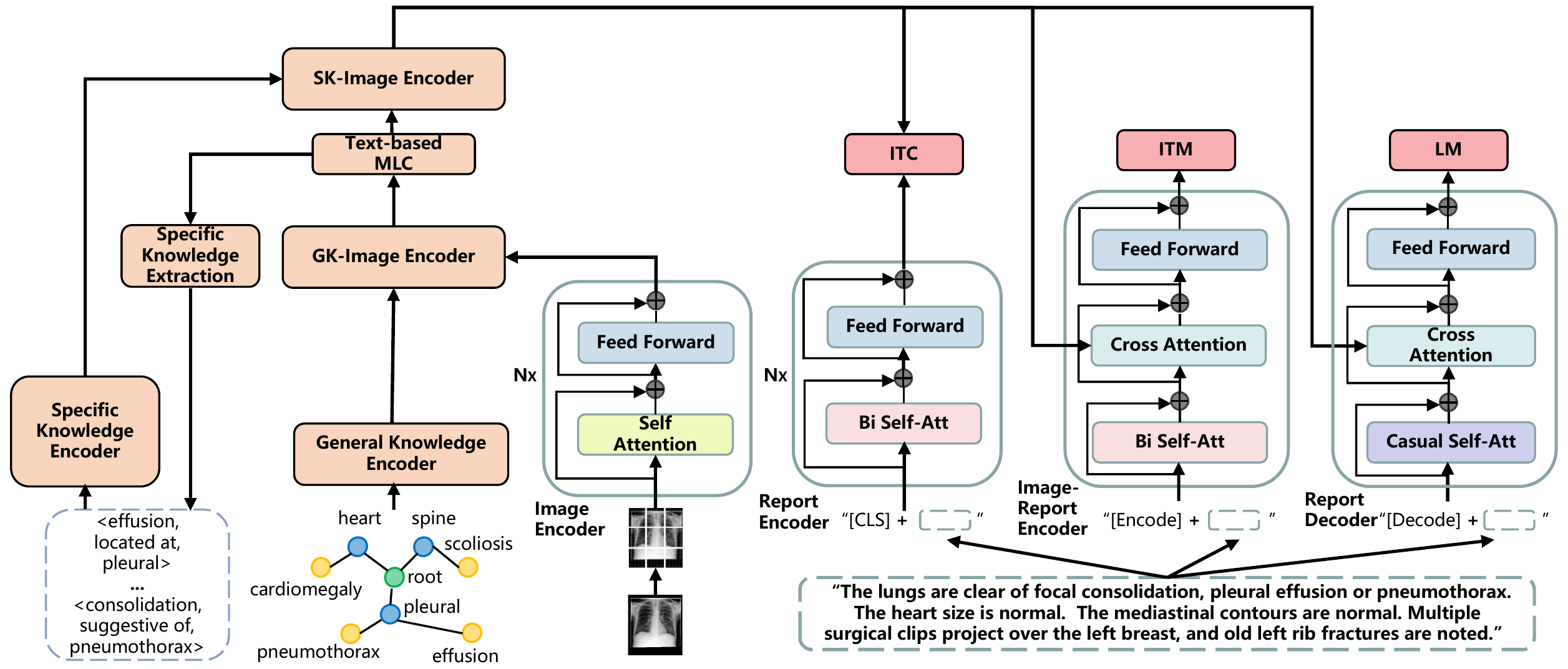}
\par\end{centering}
\caption{\label{fig:model}The overview of model architecture,   knowledge injection pipeline, and pretraining objectives of MOTOR. MOTOR contains four unimodal encoders (an image encoder, a report encoder, a general knowledge encoder, and a specific knowledge encoder), three cross-modal encoders (an image-report encoder, a GK-image encoder, and a SK-image encoder), and a report decoder. Three basic pretraining objectives, i.e., Image-Text Contrastive (ITC), Image-Text Matching (ITM), and Language Modeling (LM) are adopted for MOTOR. Both general knowledge and specific knowledge are injected into MOTOR in a complementary way with an additional text-based multi-label classification (MLC) pretraining objective.
}
\vspace{-0.4cm}
\end{figure*}

\section*{Discussion}
Medical artificial general intelligence (MAGI) can enable varying combinations of multiple medical modalities, learn new medical tasks on the fly,  and share medical knowledge sufficiently. 
To take a step towards MAGI, we have developed a new pretraining paradigm, called Medical-knOwledge-enhanced mulTimOdal pretRaining (MOTOR). 
In MOTOR, we inject the comprehensive basic medical knowledge during pretraining to build the foundation pretrained model. Then, through simple model modification and  finetuning, MOTOR can easily adapt to various downstream medical tasks.
MOTOR unifies the understanding and  generation capability in a single medical foundation model, thus can handle diverse medical tasks such as medical report generation, medical visual question answering, diagnosis classification, etc.
To enable a sufficient evaluation of the developed  model as well as provide a helpful testbed for facilitating further research, we have constructed a medical multimodal benchmark containing both understanding and generation tasks. 
Extensive experiments on various downstream tasks in our benchmark show the superiority of MOTOR over task-specific state-of-the-art models.
Moreover, we compare  MOTOR with its ``w/o knowledge'' variant on a wide range of downstream medical tasks. 
The comparison results show that MOTOR significantly outperforms this variant on all downstream tasks.
Our visualization further reveals the excellent interpretability of MOTOR, that is, the introduced medical knowledge can highlight crucial information in the redundant medical data. 

We believe that the solid step we take towards MAGI would have a broad impact on the AI-aided medical domain. 
First, MOTOR can be used to explore new modalities and novel medical tasks through simple model modification and task-oriented finetuning, due to its strong generalization ability, rich basic medical knowledge, and powerful multi-modal understanding ability.
Second, MOTOR provides a meaningful reference for developing more effective and efficient knowledge-enhanced pretraining paradigms in the medical domain. 
The complementary characteristics of different kinds of medical knowledge for assisting pretraining have been deeply investigated in MOTOR. 
Third, MOTOR is powerful in solving diverse medical tasks, which is validated by our constructed medical multi-modal benchmark. Our benchmark can also serve as a comprehensive evaluation testbed for verifying the generalization ability of other pretrained medical models. 

Despite the above advantages, MOTOR still faces potential limitations and challenges: 
1) The gap between pretraining and downstream medical tasks: Due to different task objectives between pretraining and downstream medical tasks, simple finetuning and model architecture modification for the pretrained model might lead to limited generalization ability on the downstream tasks. 
2) Limited scale of existing paired image-report data: The pretraining of MOTOR relies on paired medical image-text data, which are orders of magnitude below available single-modality medical data. The limited scale of the pretraining dataset would largely hinder the capability of the foundation medical model.
3) Complex medical imagining modalities: In the medical domain, there exist many kinds of imaging modalities, such as CT, MRI, ultrasound, etc. A general AI-aided medical system is desired to be capable of dealing with the medical images of different imaging modalities. However, it is difficult to build a comprehensive testbed containing all kinds of medical imagining modalities. Therefore, MOTOR is only evaluated on part of medical imagining modalities in the current work.

In summary, our MOTOR makes a successful attempt in realizing MAGI by imitating human experience and provides a valuable reference for further research in developing general AI-aided medical systems. There is still much room to design more powerful medical foundation models for solving various complicated medical tasks. 
For example, to  better activate the generalization ability of the foundation model, it is effective to mitigate the gap between pretraining and downstream tasks. A feasible way is to modify the training objective of downstream medical tasks to be consistent with that of pretraining, which has been shown to be effective in both the natural language processing domain and the natural image domain~\cite{Radford2021LearningTV,Song2022CLIPMA,Liu2021PretrainPA,Liu2021UnifiedMP}. 
Moreover, as suggested in~\cite{Wang2022MedCLIPCL}, to scale the usable pretraining data with low cost, it is promising to generate the paired medical image-text data from large amounts of single medical images and texts. 
Lastly, it is important to improve the visual information addressing capability of the foundation model for supporting multiple imagining modalities.

\section*{Methods}

\noindent\textbf{Method overview.}
We build MOTOR upon a recently proposed multi-modal pretraining model in the natural image domain, called BLIP~\cite{Li2022BLIPBL}. 
The overview of the model architecture,   the knowledge injection pipeline, and the pretraining objectives of MOTOR is presented in Figure~\ref{fig:model}.
Specifically, MOTOR contains four unimodal encoders, i.e., an image encoder $E_{v}$, a report encoder $E_{t}$, a general knowledge encoder $E_{gk}$, and a specific knowledge encoder $E_{sk}$. We utilize $E_{v}$, $E_{t}$, $E_{gk}$, and $E_{sk}$ to obtain the image feature $\mathbf{f}_{v}$, the report feature $\mathbf{f}_{t}$, the general knowledge feature $\mathbf{f}_{gk}$, and the specific knowledge feature $\mathbf{f}_{sk}$, respectively.
Three cross-modal encoders, i.e., an image-report encoder $E_{v\rightarrow{t}}$, a GK-image encoder $E_{gk\rightarrow{v}}$, and a SK-image encoder $E_{sk\rightarrow{v}}$, are constructed in MOTOR to enable the interactions between different modalities. ``GK'' and ``SK'' means general knowledge and specific knowledge, respectively.
A report decoder $D$ is employed for the report generation. 


To inject both general and specific knowledge in a complementary manner, we first use the general knowledge feature $\mathbf{f}_{gk}$ to enhance the image feature $\mathbf{f}_{v}$ and obtain the enhanced image feature $\mathbf{f}_{v}^{gk}$. Then, $\mathbf{f}_{v}^{gk}$ is used to retrieve similar reports from a report queue $Q$ 
 and obtain the specific knowledge feature $\mathbf{f}_{sk}$ from the retrieved reports. The specific knowledge feature $\mathbf{f}_{sk}$ further enhances the image feature $\mathbf{f}_{v}^{gk}$ to obtain the final image feature $\mathbf{f}_{v}^{gk\rightarrow{sk}}$. Enhanced by the general knowledge feature $\mathbf{f}_{gk}$ and the specific knowledge feature $\mathbf{f}_{sk}$, the image feature $\mathbf{f}_{v}^{gk\rightarrow{sk}}$ can learn both hierarchy correspondence and semantic relationships among different organs and diseases. And $\mathbf{f}_{v}^{gk\rightarrow{sk}}$ is used for calculating the pretraining objectives. 

We utilize the Image-Text Contrastive (ITC), Image-Text Matching (ITM), and Language Modeling (LM) as the basic pretraining objectives following BLIP~\cite{Li2022BLIPBL}.  
Based on the disease label provided in the pretraining image-report data, we introduce an additional pretraining objective, i.e., text-based Multi-Label Classification (MLC), to bridge the injection between the general knowledge and the specific knowledge, as well as improve the quality of the injected knowledge.
After pretraining with ITC, ITM, LM as well as our introduced text-based MLC, we use the foundation model to solve the downstream medical tasks in our benchmark through simple task-oriented adaptation.

In the following, we first describe the model architectures of unimodal encoders, cross-modal encoders, and the report decoder. Then we present the knowledge encoding and injection pipeline with the introduced text-based MLC pretraing objective. Thirdly, we give the basic pretraining objectives and the total objective function of MOTOR. Fourthly, we describe the task-oriented adaptation scheme for different downstream medical tasks. Finally, we present the dataset details and implementation details.


\noindent\textbf{Unimodal encoders.} MOTOR contains two kinds of unimodal encoders, i.e., the image encoder $E_{v}$, which is a ViT~\cite{Dosovitskiy2021AnII} model,  and the text encoder (including the report encoder $E_{t}$, the general knowledge encoder $E_{gk}$, and the specific knowledge encoder $E_{sk}$), which has the same architecture as BERT~\cite{Devlin2019BERTPO}. 
In MOTOR, both the general and specific knowledge are represented in the form of text like the report to facilitate the image-text alignment.
The concrete encoding way of knowledge are described in the ``knowledge encoding'' part of this section. 

\noindent\textbf{Cross-modal encoders.} 
MOTOR includes three cross-modal encoders, i.e., an image-report encoder $E_{v\rightarrow{t}}$, a  GK-image encoder $E_{gk\rightarrow{v}}$, and a SK-image encoder $E_{sk\rightarrow{v}}$. $E_{v\rightarrow{t}}$, $E_{gk\rightarrow{v}}$, and $E_{sk\rightarrow{v}}$ enables the interaction between images and reports, general knowledge and images, and specific knowledge and images, respectively.
The cross-modal encoder contains multiple transformer blocks which are composed of the self-attention layer, the cross-attention layer, and the feed forward network. When the report is fed to the image-report encoder $E_{v\rightarrow{t}}$, a [Encode] token is appended in the beginning. And the output embedding of [Encode] is viewed as the multimodal representation of the image-report pair.

\noindent\textbf{Report decoder.} The report decoder $D$ shares the similar architecture to the cross-modal encoders.  Different from the cross-modal encoder that use bi-directional masks in the self-attention layers, the report decoder $D$ adopts causal masks in its self-attention layers. 

\noindent\textbf{Knowledge encoding.}
Expert medical knowledge has shown great potential in improving the performance and enhancing the  explainability in many medical cross-modal tasks~\cite{li2019knowledge,li2020auxiliary,liu2020k,he2019integrating,Yang2021KnowledgeMR,Yan2022AttributedAG,Yuan2021ImprovingBP}. Moreover, some recent works have shown that injecting the knowledge into multimodal pretraining can enable the foundation model to learn better cross-modal alignment and therefore benefits downstream tasks~\cite{Cui2021ROSITAEV,yu2021ernie,Pan2022ContrastiveLP,Shen2022KLITELT}. In MOTOR, we introduce two kinds of medical knowledge, which we call general knowledge and specific knowledge, to boost the medical multi-modal pretraining. As a result,  the foundation model can learn more accurate cross-modal alignment from medical multi-modal data which has severe redundancy.  

Concretely, the general knowledge $G$ is formed by a symbolic knowledge graph. Since our pretraining dataset MIMIC-CXR consists of chest x-ray images and reports, we construct $G$ based on the chest knowledge graph proposed in~\cite{Zhang2020WhenRR,liu2021exploring}. 
$G$ contains one root node, 7 organ/tissue nodes, and 20 finding  (normal or disease keywords) nodes. The root node connects to all other nodes and  different organ/tissue nodes are connected together. Finding nodes linked to the same organ/tissue are connected to each other. We use an adjacency matrix $A=\{e_{ij}\}$ to represent the connection relationship above. When the source node $n_{i}$ connects to the target node $n_{j}$, $e_{ij}$ is set to 1. 

To obtain the general knowledge feature $\mathbf{f}_{gk}$, we first calculate the initialized node embedding $\mathbf{e}_{n}$ for the nodes in $G$ using the category word embedding $\mathbf{e}_{w}$ and the structure embedding $\mathbf{e}_{l}$:
\begin{equation}
\mathbf{e}_{n} = \mathbf{e}_{w} + \mathbf{e}_{l}.
\end{equation}
The structure embedding indicates that a node is the root, an organ, or a finding. Then we use the general knowledge encoder $E_{gk}$ to obtain $\mathbf{f}_{gk}$ based on the adjacency matrix $A$:
\begin{equation}
\mathbf{f}_{gk}=E_{gk}(\mathbf{e}_{n}, A),
\end{equation}
where $A$ serves as the visible mask~\cite{liu2020k,li2022cross} in the self-attention calculation in $E_{gk}$.

The form of the specific knowledge is a triplet containing the source entity, target entity, and their semantic relation. For image $I$, we first retrieve top $k$ similar reports $\{T_{i}\}_{i=1}^{k}$ from a report queue $Q$ based on the similarity between $I$ and $\{T_{i}\}_{i=1}^{n_{Q}}$, where $n_{Q}$ is the size of $Q$. Then we extract the named entities $E_{i}=\{e_{i}^{1},...,e_{i}^{n_{i}}\}$ ($n_{i}$ is the number of the named entities) using a named entity recognizer provided by~\cite{Qi2020StanzaAP} from each retrieved text $T_{i}$.  We query specific knowledge using the extracted entities $\{E_{i}\}_{i=1}^{k}$ from the public available knowledge graph RadGraph~\cite{jain2021radgraph}, which contains plenty of clinic radiology entities with their semantic relations, such as $<$consolidation, suggestive of, pneumothorax$>$ and $<$effusion, located at, pleural$>$. Denote the specific knowledge set of image $I$ as $K_{I}=\{K_{I}^{j}\}_{j=1}^{n_{K}}$, where $K_{I}^{j}$ is the $j$-th triplet and $n_{K}$ is the size of $K_{I}$.

\begin{figure*}
\begin{centering}
\includegraphics[width=0.98\linewidth]{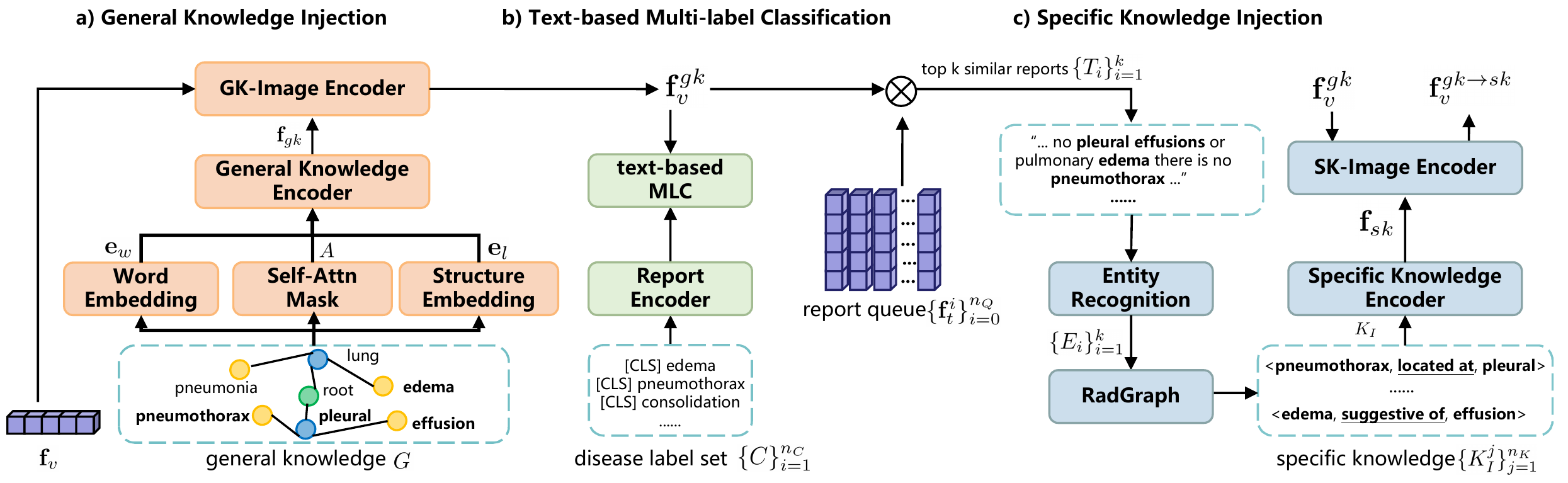}
\par\end{centering}
\caption{\label{fig:knowledge}The illustration of the knowledge injection pipeline. During a) general knowledge injection, the general knowledge $G$ is fed to the general knowledge encoder $E_{gk}$ to obtain the general knowledge feature $\mathbf{f}_{gk}$. And the image feature $\mathbf{f}_{v}$ is enhanced by $\mathbf{f}_{gk}$ through the GK-image encoder $E_{gk\rightarrow v}$ to obtain $\mathbf{f}_{v}^{gk}$. Then in b)  text-based MLC,  $\mathbf{f}_{v}^{gk}$ is used for multi-label classification based on the disease label set $\{C\}_{i=1}^{n_{C}}$ represented in the textual form. Finally in c) specific knowledge injection, specific knowledge $K_{I}=\{K_{I}^{j}\}_{j=1}^{n_{K}}$ is extracted from RadGraph~\cite{jain2021radgraph} based on $\mathbf{f}_{v}^{gk}$ and fed to the specific knowledge encoder $E_{sk}$ to obtain the specific knowledge feature $\mathbf{f}_{sk}$. The image feature $\mathbf{f}_{v}^{gk}$ is further enhanced by $\mathbf{f}_{sk}$ through the SK-image encoder $E_{sk\rightarrow v}$ to generate  $\mathbf{f}_{v}^{gk\rightarrow sk}$, which are used for calculating basic pretraining objectives.
}
\vspace{-0.4cm}
\end{figure*}


The specific knowledge feature $\mathbf{f}_{sk}$ is obtained by the specific knowledge  encoder $E_{sk}$. 
Concretely, we concatenate the triplets $\{K_{I}^{j}\}_{j=1}^{n_{K}}$ in $K_{I}$ into one single sentence and feed the sentence to $E_{sk}$ to obtain $\mathbf{f}_{sk}$:
\begin{equation}
\label{specific knowledge encoding}
\mathbf{f}_{sk}=E_{sk}([k_{I}^{1};...;k_{I}^{n_{K}}]).
\end{equation}

\noindent\textbf{Knowledge injection.}
Before the knowledge injection, we use the image encoder $E_{v}$ to obtain the image feature $\mathbf{f}_{v}$ of image $I$:
\begin{equation}
\mathbf{f}_{v}=E_{v}(I).
\end{equation}
Then we inject the knowledge to enhance $\mathbf{f}_{v}$ for improving the cross-modal alignment. The pipeline of the knowledge injection is given in Figure~\ref{fig:knowledge}. Specifically, we first inject the general knowledge $\mathbf{f}_{gk}$ for $\mathbf{f}_{v}$ to generate the enhanced image feature $\mathbf{f}_{v}^{gk}$, as shown in Figure~\ref{fig:knowledge}a). Then we introduce an additional text-based multi-label classification (MLC) pretraining task for $\mathbf{f}_{v}^{gk}$ to improve the quality of the subsequent specific knowledge injection, as shown in Figure~\ref{fig:knowledge}b). Finally, we inject the specific knowledge $\mathbf{f}_{sk}$ for $\mathbf{f}_{v}^{gk}$ to generate the enhanced image feature $\mathbf{f}_{v}^{gk\rightarrow sk}$, as shown in Figure~\ref{fig:knowledge}c). The  enhanced feature $\mathbf{f}_{v}^{gk\rightarrow sk}$ is finally utilized for calculating basic pretraining objectives. 

The general knowledge injection is performed through the GK-image encoder $E_{gk\rightarrow{v}}$.
Specifically, the general knowledge feature $\mathbf{f}_{gk}$ enhances the image feature $\mathbf{f}_{v}$ by:
\begin{equation}
\mathbf{f}_{v}^{gk}=E_{gk\rightarrow{v}}(\mathbf{f}_{v}, \mathbf{f}_{gk}, \mathbf{f}_{gk}),
\end{equation}
where $\mathbf{f}_{v}$, $\mathbf{f}_{gk}$, $\mathbf{f}_{gk}$ serve as the query, the key, and the value, respectively, in the cross attention calculation in $E_{gk\rightarrow{v}}$.
Enhanced by $\mathbf{f}_{gk}$, the updated image feature $\mathbf{f}_{v}^{gk}$ can capture the global relations among different organs and diseases. 

To inject the specific knowledge, we first construct a first-in-first-out report queue $Q$ using the reports in the pretraining dataset. Denote the report features in $Q$ as $\{\mathbf{f}_{t}^{i}\}_{i=0}^{n_{Q}}$, where $n_{Q}$ is the size of $Q$. We calculate the similarity between the image feature $\mathbf{f}_{v}^{gk}$ and the report features $\{\mathbf{f}_{t}^{i}\}_{i=0}^{n_{Q}}$, and retrieve top $k$ similar reports $\{T_{i}\}_{i=1}^{k}$. After extracting the specific knowledge $K_{I}$ from $\{T_{i}\}_{i=1}^{k}$, we get the specific knowledge feature $\mathbf{f}_{sk}$ by feeding $K_{I}$ into the text encoder $E_{sk}$ (Eq.~\ref{specific knowledge encoding}). Then we utilize $\mathbf{f}_{sk}$ to enhance $\mathbf{f}_{v}^{gk}$ through the SK-image encoder $E_{sk\rightarrow{v}}$:
\begin{equation}
\mathbf{f}_{v}^{gk\rightarrow{sk}}=E_{sk\rightarrow{v}}(\mathbf{f}_{v}^{gk}, \mathbf{f}_{sk}, \mathbf{f}_{sk}),
\end{equation}
where $\mathbf{f}_{v}^{gk}$, $\mathbf{f}_{sk}$, $\mathbf{f}_{sk}$ serve as the query, the key, and the value, respectively, in the cross attention calculation in $E_{sk\rightarrow{v}}$.
The enhanced image feature $\mathbf{f}_{v}^{gk\rightarrow{sk}}$ is finally used for calculating the pretraining objectives of ITC, ITM, and LM.
As shown in Figure~\ref{fig:knowledge}, the general knowledge and the specific knowledge  help the model learn the hierarchy and semantic relations among organs and diseases, respectively, which effectively enhances the image feature in a complementary manner for facilitating the modality alignment.

To better bridge the injection between the general knowledge and the specific knowledge, we introduce an additional text-based MLC pretraining task for the image feature $\mathbf{f}_{v}^{gk}$. The text-based MLC pretraining task aligns $\mathbf{f}_{v}^{gk}$ to the corresponding disease labels represented in the textual form.   As a result, the image feature $\mathbf{f}_{v}^{gk}$ is capable of capturing instance-related entities in the report. Therefore, the report retrieval accuracy in the specific knowledge injection process based on $\mathbf{f}_{v}^{gk}$ can be largely improved to ensure the quality of the injected specific knowledge.
Concretely, we first extract the disease labels~\cite{Irvin2019CheXpertAL} in the pretraining dataset MIMIC-CXR~\cite{Johnson2019MIMICCXRAL} to build the label set $\{C\}_{i=1}^{n_{C}}$, where $n_{C}=14$. Then we represent each disease label $C_{i}$ as the textual label feature $\mathbf{f}_{C_{i}}$ using the report encoder $E_{t}$:
\begin{equation}
\mathbf{f}_{C_{i}}=E_{t}(C_{i}).
\end{equation}
A [CLS] token is added to $C_{i}$ when obtaining $\mathbf{f}_{C_{i}}$. We calculate the similarity $p_{C_{i}}$ between the updated image feature $\mathbf{f}_{v}^{gk}$ and each label feature $\mathbf{f}_{C_{i}}$ by:
\begin{equation}
p_{C_{i}}=\mathbf{f}_{v}^{gk} \cdot \mathbf{f}_{C_{i}}.
\end{equation}
Then $\mathbf{p}_{C}=\{p_{C_{i}}\}_{i=1}^{n_{C}}$ is used to calculate the MLC loss $\mathcal{L}_{mlc}$ based on the ground-truth disease label $y_{I}$ of the image $I$:
\begin{equation}
\mathcal{L}_{\mathrm{mlc}}=-\mathbb{E}_{I\sim D}y_{I}\mathrm{log}(\mathbf{p}_{C}).
\end{equation}
$\mathcal{L}_{mlc}$ is a binary cross-entropy loss. Note that by capturing the global relations among different organs and diseases through the general knowledge $G$, the multi-label classification performance based on the updated feature $\mathbf{f}_{v}^{gk}$ can also be effective enhanced. 
Consequently, the image feature $\mathbf{f}_{v}^{gk}$ can be better aligned to correct clinical entities to improve the quality of the specific knowledge injection. 


\noindent\textbf{Pretraining objectives.}
Besides the text-based MLC task described above, we use three basic pretext tasks similar to~\cite{Li2022BLIPBL} during pretraining. Specifically, the Image-Text Contrastive (ITC) task is adopted for improving the image encoder $E_{v}$ and the report encoder $E_{t}$ by enforcing the alignment of positive image-report pairs in the feature space. The Image-Text Matching (ITM) task is designed for the image-report encoder $E_{v\rightarrow{t}}$ by requiring the model to predict whether an image-report pair is positive or negative given the related multi-modal features. The Language Modeling (LM) task is conducted for activating the report decoder $D$ by asking the model to generate reports given an image through optimizing a cross-entropy loss. 

Similar to~\cite{He2019MomentumCF}, we keep two queues to store the most recent $M$ image-report features from the momentum image and text encoders. For ITC, we first calculate the softmax-normalized image-to-report similarity $f_m^{\mathrm{i2t}}(I)$ and the report-to-image similarity $f_m^{\mathrm{t2i}}(T)$ for the image $I$ and its paired report $T$ by:
\begin{equation}
\begin{aligned}
    f_m^{\mathrm{i2t}}(I) &=  \frac{\mathrm{exp}(s(I,T_{m})/\tau)}{\textstyle \sum_{m=1}^{M}\mathrm{exp}(s(I,T_{m})/\tau)},\\
    f_m^{\mathrm{t2i}}(T) &=  \frac{\mathrm{exp}(s(T,I_{m})/\tau)}{\textstyle \sum_{m=1}^{M}\mathrm{exp}(s(T,I_{m})/\tau)}, 
    \end{aligned}
\end{equation}
where ${\tau}$ is the temperature parameter and $s(I,T)$ denotes the similarity between the unimodal features of $I$ and $T$. The unimodal feature of $I$ is the enhanced image feature $\mathbf{f}_{v}^{gk\rightarrow sk}$, and the unimodal feature of $I_{m}$ is also the enhanced image feature obtained in a similar way to $\mathbf{f}_{v}^{gk\rightarrow sk}$. Then the ground-truth similarities $\mathbf{g}^{\mathrm{i2t}}(I)$ and $\mathbf{g}^{\mathrm{t2i}}(T)$ are utilized to calculate the ITC loss $\mathcal{L}_{itc}$:
\begin{equation}
\begin{aligned}
\mathcal{L}_{\mathrm{itc}}=&\frac{1}{2}\mathbb{E}_{(I,T)\sim D}[-(\mathbf{g}^{\mathrm{i2t}}(I)\mathrm{log}(\mathbf{f}^{\mathrm{i2t}}(I))\\
&+
\mathbf{g}^{\mathrm{t2i}}(T)\mathrm{log}(\mathbf{f}^{t2i}(T)))].
\end{aligned}
\end{equation}

ITM is a binary classification task.
Concretely, the image-report encoder $E_{v\rightarrow t}$ first receives the enhanced image feature $\mathbf{f}_{v}^{gk\rightarrow sk}$ of the image $I$ and the report feature $\mathbf{f}_{T}$ to produce the multi-modal feature $\mathbf{\tilde{f}}_{T}$:
\begin{equation}
\mathbf{\tilde{f}}_{T}=E_{v\rightarrow t}(\mathbf{f}_{T},\mathbf{f}_{v}^{gk\rightarrow sk},\mathbf{f}_{v}^{gk\rightarrow sk}),
\end{equation}
where $\mathbf{f}_{T}$, $\mathbf{f}_{v}^{gk\rightarrow sk}$, $\mathbf{f}_{v}^{gk\rightarrow sk}$ serve as the query, the key, and the value, respectively, in the cross attention calculation in $E_{v\rightarrow t}$.
Then the feature of [Encode] (See the ``Cross-modal encoder'' part) is extracted from $\mathbf{\tilde{f}}_{T}$ and fed into a linear layer to produce a two-class prediction probability $p^{\mathrm{itm}}(I,T)$. The probability $p^{\mathrm{itm}}(I,T)$ indicates whether the current image and text match (or not). Based on $p^{\mathrm{itm}}(I,T)$ and the 2-dimensional ground-truth label ${g^{\mathrm{itm}}}$, the ITM loss $\mathcal{L}_{\mathrm{itm}}$ is calculated by: 
\begin{equation}
        \mathcal{L}_{\mathrm{itm}}=-\mathbb{E}_{(I,T)\sim D}g^{\mathrm{itm}}\mathrm{log}(p^{\mathrm{itm}}(I,T)).
\end{equation}

In LM, the report is generated in an auto-regressive manner, and the LM loss $\mathcal{L}_{\mathrm{lm}}$  is calculated  based on the enhanced image feature $\mathbf{f}_{v}^{gk\rightarrow sk}$ and the ground-truth report $T=\{y_{1},y_{2},...,y_{n_{T}}\}$:
\begin{equation}
        \mathcal{L}_{\mathrm{lm}}=-\sum_{i=1}^{n_{T}}\mathrm{log}(p(y_{i}|y_{1:i-1}, \mathbf{f}_{v}^{gk\rightarrow sk})),
\end{equation}
where $p(y_{i}|y_{1:i-1}, \mathbf{f}_{v}^{gk\rightarrow sk})$ is generated via the report decoder $D$ and $n_{T}$ is the number of words. With the MLC loss $\mathcal{L}_{\mathrm{mlc}}$, the final objective  $\mathcal{L}$ of MOTOR is calculated by:

\begin{equation}
        \mathcal{L}=\mathcal{L}_{\mathrm{itc}}+\lambda_{1}\mathcal{L}_{\mathrm{itm}}+\lambda_{2}\mathcal{L}_{\mathrm{lm}}+\lambda_{3}\mathcal{L}_{\mathrm{mlc}},
\end{equation}
where $\lambda_{1}$, $\lambda_{2}$, and $\lambda_{3}$ are the balance factors.

\noindent\textbf{Downstream Task Adaptation.}
After pretraining, we use the pretrained foundation model for conducting  downstream medical tasks through task-oriented adaptation.
For the image-report retrieval task, we directly finetune the  image encoder $E_{v}$ and report encoder $E_{t}$ in the foundation model with the image-text contrastive (ITC) loss $\mathcal{L}_{\mathrm{itc}}$ and the image-text matching (ITM) loss  $\mathcal{L}_{\mathrm{itm}}$. For the medical report generation task, we finetune the  image encoder $E_{v}$ and  report decoder $D$ in the foundation model using the language modeling (LM) loss $\mathcal{L}_{\mathrm{lm}}$. For the diagnosis classification task, we simply add 14 linear classifier heads on top of the image encoder $E_{v}$  in the foundation model. We train the whole classification model using the binary cross-entropy loss to perform multi-label classification based on the disease labels provided in the diagnosis classification datasets. For the medical visual question answering (VQA) task, we finetune the  image encoder $E_{v}$ and image-report encoder $E_{v\rightarrow t}$ with the cross-entropy loss~\cite{Zhan2020MedicalVQ} using the ground-truth answer labels in the medical VQA task datasets. 

\noindent\textbf{Dataset details.}
\textbf{MIMIC-CXR}~\cite{Johnson2019MIMICCXRAL} is a publicly available chest radiograph dataset, containing 77,110 chest X-ray images and 227,835 reports.
The dataset is split into 368,960 images/222,758 reports for training, 2,991 images/1,808 reports for validation, and 5,159 images/3,269
reports for testing.

\textbf{IU-Xray}~\cite{DemnerFushman2016PreparingAC} is a publicly available X-ray chest dataset published by Indiana University. IU-Xray contains 7,470 chest X-ray images and 3,955 medical reports, among which the training set contains 6,470 images, the validation set contains 500 images, and the test set contains 500 images. 

\textbf{ChestX-ray14}~\cite{Wang2017ChestXRay8HC} is a medical imaging dataset containing 112,120 front-view X-ray images of 30,805 unique patients with 14 common diseases labelled.

\textbf{VQA-RAD}~\cite{Lau2018ADO} is a manually constructed dataset. Researchers collect radiographic images and their corresponding question-answer pairs provided by clinicians to build the dataset. VQA-RAD contains 315 images and 3,515 questions, of which 1,515 are open-ended and 2,000 are closed-ended. On average, there are 10 corresponding questions per image.

\textbf{SLAKE}~\cite{Liu2021SlakeAS} is a semantically annotated, and knowledge-enhanced bilingual dataset. SLAKE contains 642 radiographic images, covering 12 diseases and 39 organs. Dividing into 10 types (e.g., color, position), 14,028 questions are split into the training set, validation set and test set accounting for 70\%, 15\%, 15\% respectively.

\noindent\textbf{Implementation details.}
The MIMIC-CXR~\cite{Johnson2019MIMICCXRAL} dataset contains both the frontal and the lateral view images, and we use the same image encoder for different views following~\cite{Chen2020GeneratingRR}. Considering the large gap between medical texts and general texts, we use SciBERT~\cite{Beltagy2019SciBERTAP} instead of BERT~\cite{Devlin2019BERTPO} to serve as the text encoder. 
We pretrain MOTOR on 8 NVIDIA V100 GPUs with the batch size 32 and 30 epochs. The learning rate is set as 1e-4 and the optimizer is 
AdamW. During the specific knowledge injection, we retrieve the top 3 similar texts from the report queue $Q$, i.e., $k$ is set to 3. And $n_{Q}$ is set to 65536. The max length of the specific knowledge sentence is set to 90. The loss weights $\lambda_{1}$, $\lambda_{2}$, and $\lambda_{3}$ are set to 1 empirically.


In the image-report retrieval task, we finetune the image encoder and the report encoder using the AdamW optimizer. The batch size, the total training epochs, and the learning rate are set to 32, 30, and 5e-5, respectively.
In the medical report generation task, we finetune the image encoder and the report decoder for 50 epochs with the AdamW optimizer following~\cite{Chen2020GeneratingRR}. The batch size is 32 and the learning rate is 1e-5. 
For the diagnosis classification task, the learning rate, the batch size, and the training epochs on both MIMIC-CXR~\cite{Johnson2019MIMICCXRAL} and ChestX-ray14~\cite{Wang2017ChestXRay8HC} are set as 1e-5, 32, and 50, respectively.
For the medical visual question answering task, we first use the pretrained task type classifier for distinguishing the answers of different types, i.e., open-ended or closed-ended following~\cite{Zhan2020MedicalVQ}. Then we use two image-report encoders whose architectures are the same for separately addressing the questions whose answers are open-ended and closed-ended. We train the model with 200 epochs and 300 epochs on VQA-RAD~\cite{Lau2018ADO} and SLAKE~\cite{Liu2021SlakeAS}, respectively. The learning rate is set as 5e-3 and the optimizer is AdamW. 
\section*{Data availability}
All datasets used in this work are publicly available. The MIMIC-CXR dataset is available at \url{https://physionet.org/content/mimic-cxr/2.0.0/}. The IU-Xray dataset is available at \url{https://openi.nlm.nih.gov/}. The ChestX-ray14 dataset is available at \url{https://nihcc.app.box.com/v/ChestXray-NIHCC}. The VQA-RAD dataset is available at \url{https://www.nature.com/articles/sdata2018251#data-citations}. The SLAKE dataset is available at \url{https://www.med-vqa.com/slake/}.  

\section*{Code availability}
The code is available at \url{https://github.com/chenzcv7/MOTOR}.


\section*{Competing interests}
The authors declare no competing interests.

\section*{Author contributions}
B.L. and M.L. contributed to the original idea, model design, and experimental analysis. B.L., M.L., and Z.C. wrote the manuscript. Z.C., H.L., M.L., and B.L. conducted the experiments. W.C. and L.Y. conducted the human study. The review and editing of the manuscript were carried out by H.X., Y.Z., J.L., W.C., L.Y., S.Z., C.W., L.C., X.C, Y.Y. and L.X. The entire project was supervised by X.L.


%






\ifCLASSOPTIONcaptionsoff
  \newpage
\fi



%



\bibliographystyle{IEEEtran}
    \bibliography{IEEEabrv,egbib}

%








\end{document}